# DeepSEED: 3D Squeeze-and-Excitation Encoder-Decoder Convolutional Neural Networks for Pulmonary Nodule Detection

*Yuemeng Li, Yong Fan*

Center for Biomedical Image Computing and Analytics, Department of Radiology, Perelman School of Medicine, University of Pennsylvania, Philadelphia, PA, 19104

**ABSTRACT**

Pulmonary nodule detection plays an important role in lung cancer screening with low-dose computed tomography (CT) scans. It remains challenging to build nodule detection deep learning models with good generalization performance due to unbalanced positive and negative samples. In order to overcome this problem and further improve state-of-the-art nodule detection methods, we develop a novel deep 3D convolutional neural network with an Encoder-Decoder structure in conjunction with a region proposal network. Particularly, we utilize a dynamically scaled cross entropy loss to reduce the false positive rate and combat the sample imbalance problem associated with nodule detection. We adopt the squeeze-and-excitation structure to learn effective image features and utilize inter-dependency information of different feature maps. We have validated our method based on publicly available CT scans with manually labelled ground-truth obtained from LIDC/IDRI dataset and its subset LUNA16 with thinner slices. Ablation studies and experimental results have demonstrated that our method could outperform state-of-the-art nodule detection methods by a large margin.

## 1. INTRODUCTION

Lung cancer is one of the most commonly diagnosed cancers in the united states and worldwide [1]. Lung cancer screening with low-dose computed tomography (CT) scans is an effective way to reduce mortality of lung cancer, which relies on accurate pulmonary nodule detection. To achieve accurate and effective pulmonary nodule detection in low-dose CT scans, a number of deep learning based computer-aid detection (CAD) methods have been developed based on convolutional neural networks (CNNs) [2, 3, 4].

Although the deep learning based methods approach the pulmonary nodule detection problem in many different ways [5, 4, 6, 7, 3, 2], most of them adopt a two-stage strategy [8], consisting of 1) nodule candidate detection to identify nodule regions with reduced false positive (FP) in a classification setting; and 2) nodule bounding boxes prediction to estimate spatial locations of positive nodules in a regression setting. For instance, 2D CNNs are adopted to build pulmonary nodule detection models on multiple views of 2D CT slices [5], and 3D CNNs are trained with an online sample filtering scheme for candidate screening in conjunction with a residual network for FP reduction [6]. Due to their multi-stage frameworks, those methods are computationally expensive, and their final outputs might unfavorably depend on their early stage's performance.

Recent studies have demonstrated that efficient pulmonary nodule detection could be achieved by end-to-end learning methods. Particularly, a nodule detection network has been trained end-to-end using 3D CNNs [7]. In addition, several pulmonary nodule detection methods have been developed by adopting faster R-CNN which consists of a region proposal network (RPN) to generate region proposals and a detection network to detect objects from proposals [9]. In particular, a 2D faster R-CNN model has been built for candidate detection followed by a 3D CNN for FP reduction [4], a faster R-CNN model has been built upon a 3D RPN with an encoder-decoder structure [3], and a faster R-CNN model with dual path blocks has been built to take the advantages of residual learning and dense connection for nodule detection [2]. In all these methods, nodule candidate detection is the most challenging task due to the unbalanced sample problem, i.e., a large number of negative samples versus a relatively small number of positive samples.

To achieve improved pulmonary nodule detection in low-dose CT scans, we develop a novel deep Squeeze-and-Excitation Encoder-Decoder (DeepSEED) network to detect nodules in one single step. Particularly, DeepSEED is built upon a 3D RPN with a squeeze-and-excitation structure [10] to effectively learn image features for accurately detecting nodules, DeepSEED's RPN is built upon an Encoder Decoder structure to effectively utilized multiscale context image information, and DeepSEED's RPN utilizes a dynamically scaled cross entropy loss, namely focal loss [11], to combat the imbalance of positive and negative samples. We have evaluated our method and compared it with state-of-the-art methods based on large data sets, including LIDC/IDRI dataset and its subset LUNA16 with thinner slices. Extensive validation experimental results have demonstrated that our method could achieve better nodule detection performance than the alternative methods under comparison.

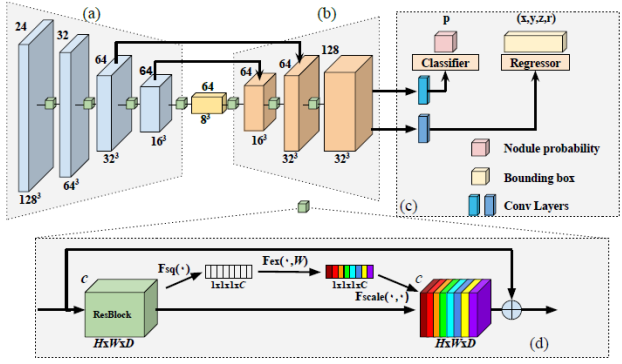

**Fig. 1**. The architecture of DeepSEED. Number on the top left of each block indicates the number of channels and number on the bottom right indicates spatial dimensions of input feature maps. (a) An encoder with a 3D ResNet-18 structure. (b) A decoder with its feature dimension expanded. (c) A region proposal network for identifying candidate nodules and predicting their bounding boxes. (d) The squeeze-and-excitation (SE) residual blocks that are adopted in all convolutional blocks in (a) and (b). Particularly, each residual block is squeezed into a $C$-dimension feature vector (white blocks) that passes through a gating mechanism to apply the spatial encoding (color blocks) to the residual block, with different colors indicating different weights.

## 2. METHODS

The proposed DeepSEED network for automatic pulmonary nodule detection in low-dose CT scans is illustrated in Fig. 1. Specifically, DeepSEED is built upon a 3D region proposal network (RPN) [9] with an Encoder-Decoder structure, enhanced by a Squeeze-and-Excitation structure [10] to fully exploit image features and a focal loss [11] to effectively identify true positives among all nodule candidates.

### 2.1. Network Structure

In this study, we adopt a 3D RPN structure as backbone due to its outstanding performance in nodule detection tasks [2, 3]. The backbone contains a ResNet-18 structure as the encoder illustrated in Fig. 1(a), and the decoder consists of 3 layers of residual blocks with skip connections as illustrated in Fig. 1(b). Each residual block contains a Squeeze-and-Excitation unit as illustrated in Fig. 1(d). The RPN's input is a 3D CT image with anchors that are candidate nodule bounding boxes. The RPN's outputs include each anchor's classification probability score $p$ to be a nodule and the nodule's spatial information including its coordinates $x, y, z$ and the nodule radius $r$. The anchor's size could be set according to the distribution of nodule sizes of the training data. Implementation details are presented in Section of Experimental Results.

We adopt a Squeeze-and-Excitation (SE) structure in our network due to its success in image classification [10]. Particularly, the SE structure could effectively utilize channel inter-dependency with an enhanced spatial encoding to make the network to adaptively adjust the weights of each feature map for the nodule detection task. As illustrated in Fig. 1(d), for each convolutional layer, we set $U = F(X)$, where $X$ is the input feature, and $U = [u_1, u_2, ..., u_c]$, $u_i \in R^{H \times W \times D}$ is the output feature matrix after convolution. In order to squeeze the global spatial information into a channel descriptor, we use channel-wise global average pooling. The output is denoted by $z_i = F_{sq}(u_i)$, which is an element of $Z \in R^C$ generated by squeezing $U$ through the spatial dimensions of $H \times W \times D$.

The channel-wise output of the squeeze operation is used to modulate inter-dependencies of all channels through excitation, a gating mechanism with a sigmoid activation:
$$s_i = F_{ex}(u_i) = \sigma(g(Z, W)) = \sigma(W_2 \delta(W_1 Z)), \quad (1)$$
where $\sigma(\cdot)$ is the sigmoid activation function, $\delta(\cdot)$ is the ReLU function [12], $W_1 \in R^{\frac{C}{r} \times C}$ contains parameters for the dimensionality-reduction layers, and $W_2 \in R^{C \times \frac{C}{r}}$ contains parameters for the dimensionality increasing layers. In the present study we set $r = 16$, a commonly used setting for SE structures [10].

A channel-wised multiplication $F_{scale}$ between each excitation scalar $s_i$ and the feature map $u_i$ is finally applied to generate the final re-scaling feature output $U$:
$$U = F_{scale}(u_i, s_i) = s_i \cdot u_i, \quad (2)$$
The output features of squeeze-and-excitation are obtained before the identity shortcut connection for each residual block.

### 2.2. Classification and Regression Loss Functions

The RPN is trained to minimize a nodule classification loss $L_{cls}$ for distinguishing positive from negative nodules and a regression loss $L_{reg}$ for predicting bounding boxes. To combat the imbalance of positive and negative samples of nodule regions and to correct misclassified samples, we adopt a focal loss [11] to compute the nodule classification loss $L_{cls}$. Given a positive sample's classification output probability $p$, the focal loss function is defined as:
$$L_{cls} = -\alpha(1 - p_t)^\gamma \log(p_t), \quad (3)$$
where $p_t = p$ if its ground truth class label $y = 1$, otherwise $p_t = 1 - p$, $\alpha$ is a balance factor for the focal loss, and $\gamma$ is a tunable focusing parameter. In the present study, $\alpha=0.5$ and $\gamma = 2$ as suggested by [11].

The regression loss is defined as:
$$L_{reg} = \sum_k S(G_k, P_k), \quad (4)$$
where $S(\cdot)$ is the smoothed L1-norm function [9], $G_k = \left(\frac{x_g - x_a}{r_a}, \frac{y_g - y_a}{r_a}, \frac{z_g - z_a}{r_a}, \log\left(\frac{r_g}{r_a}\right)\right)$ is the parameterized ground truth, $P_k = \left(\frac{x - x_a}{r_a}, \frac{y - y_a}{r_a}, \frac{z - z_a}{r_a}, \log\left(\frac{r}{r_a}\right)\right)$ is the corresponding parameterized prediction, $x_a, y_a, z_a$ and $r_a$ are spatial location and radius of an anchor under consideration respectively, and $k$ is an index of the anchors in a mini-batch.

The overall loss function is written as:
$$L = L_{cls} + p^* L_{reg}, \quad (5)$$
where $p^* = 1$ for positive samples and 0 for negative samples.

**Table 1.** FROC of different numbers of false positives per scan obtained by different methods under comparison on LUNA and LIDC/IDRI datasets. The first 3 rows in gray background show results on LUNA dataset, and the last 2 rows show results on LIDC/IDRI dataset.

| FROC | 0.125 | 0.25 | 0.5 | 1 | 2 | 4 | 8 | Mean |
|---|---|---|---|---|---|---|---|---|
| 3D RPN [3] | 0.662 | 0.746 | 0.815 | 0.864 | 0.902 | 0.918 | 0.932 | 0.834 |
| DeepLung [2] | 0.692 | 0.769 | 0.824 | 0.865 | 0.893 | 0.917 | 0.933 | 0.842 |
| DeepSeed (ours) | 0.739 | 0.803 | 0.858 | 0.888 | 0.907 | 0.916 | 0.920 | 0.862 |
| 3D RPN [3] | 0.552 | 0.630 | 0.700 | 0.751 | 0.788 | 0.823 | 0.852 | 0.728 |
| DeepSeed (ours) | 0.600 | 0.674 | 0.751 | 0.824 | 0.850 | 0.853 | 0.859 | 0.773 |

## 3. EXPERIMENTAL RESULTS

We evaluated our method based on LIDC/IDRI [13] and LUNA16 [14], and compared it with state-of-the-art nodule detection methods built upon RPNs [3, 2]. Particularly, a 3D RPN [3] was adopted as the baseline, and we further compared our method with Dual Path Network [2] due to its outstanding performance on LUNA16 dataset.

### 3.1. Datasets

We used publicly available LIDC/IDRI dataset [13] and its subset LUNA16 [14] to evaluate the nodule detection methods under comparison. Both the datasets labeled nodules greater than 3 mm. CT scans of LUNA16 and LIDC/IDRI differ mainly in their slice thickness. LUNA16 contains 888 cases of CT scans with 1186 labeled nodules, obtained from LIDC dataset by excluding CT scans with slice thickness greater than 2.5 mm. Another noticeable difference between LUNA16 and LIDC/IDRI is their nodule size distribution. The mean nodule size of LUNA16 is 8.3 mm with a variance of 4.8 mm whereas that of LIDC/IDRI is 12.8 mm with a variance of 10.6 mm. Nodule detection is more challenging on LIDC/IDRI dataset because of its lower spatial resolution. We used a 10-fold cross-validation on LUNA16 to evaluate all the methods under comparison and these models were then applied to 122 subjects from LIDC/IDRI dataset (excluding LUNA16) for testing their generalization performance.

### 3.2. Implementation Details

We preprocessed each input image with a binary segmentation mask to segment lungs, clipped pixel values to the range [-1200, 600], and finally normalized pixel values to [0, 1]. We divided images into patches of 128×128×128 as the network input for training. All the patches were randomly cropped, and additional data augmentation was implemented with random flipping and scaling with the ratio ranging 0.75 to 1.25. To generate training anchors, boxes with Intersection over Union (IoU) greater than 0.5 were defined as positive samples and smaller than 0.02 as negative samples. Anchors were generated with different sizes of [5, 10, 20]. For testing, we divided images into patches of size 208×208×208 with an overlapping of 32 pixels for neighboring patches. The model performance was evaluated using an official script of Free-Response Receiver Operating Characteristic (FROC) analysis provided by LUNA16 [14]. In the FROC analysis, sensitivity is defined as a function of the average number of false positives per scan (FPs/scan). The overall score was evaluated at 0.125, 0.25, 0.5, 1, 2, 4 and 8 false positives per scan. We trained our network on 8 NVIDIA 1080Ti GPUs with the batch size of 40 and learning rate of 0.01.

### 3.3. Performance of Different Methods under Comparison

We implemented 3D RPN [3] as our baseline and compared our method based on LUNA 16 dataset with those reported in DeepLung [2]. Table 1 summarizes performance of different methods under comparison. The proposed DeepSEED achieved the highest average FROC score of 86.2% on LUNA16 and 77.3% on LIDC/IDRI. Specifically, DeepSEED outperformed the state-of-the-art nodule detection methods by 4.1% and 3.0% on 0.125 to 2 false positives per scan respectively. The results on LIDC/IDRI dataset also indicated that our network had a better generalization performance since our network was trained on LUNA16. These results are also supported by FROC curves obtained on LIDC/IDRI dataset by the proposed and the 3D RPN models trained on LUNA 16, as shown in Fig. 2.

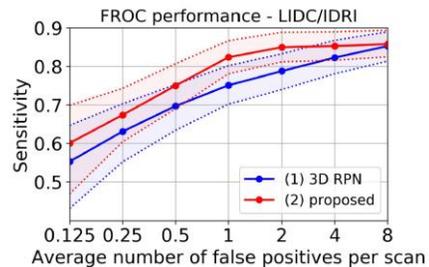

**Fig. 2**. FROC curves obtained by the 3D ResNet-18 backbone and DeepSEED on LIDC/IDRI dataset. The dash-lines show the corresponding upper and lower bounding after bootstrapping.

We also compared DeepSEED and the 3D RPN in terms of their accuracy for predicting size of bounding boxes for testing samples with correct nodule detection results. As indicated by representative results shown in Fig. 3, DeepSEED predicted precise bounding boxes sizes and accurate locations of bounding boxes, better than the 3D RPN model did.

### 3.4. Ablation studies

To investigate how different components of our method contribute to the nodule detection, we compared the 3D RPN with DeepSEED and its degraded versions without focal loss

(proposed no Focal) and without SE blocks (proposed no SE). The ablation was performed on a randomly selected fold of the cross-validation experiments. As shown in Fig. 4, the sensitivity of RPN was much lower than DeepSEED and its degraded versions. For settings of average numbers of FPs/scan smaller than or equal to 1, the degraded versions of DeepSEED obtained better FROC scores, but DeepSEED had better performance for settings of average numbers of FPs/scan greater than 1. These experimental results indicated that both the SE structure and the focal loss could improve the 3D RPN's performance if they were used alone, and their combination could further improve the 3D RPN's performance for setting of average numbers of FPs/scan greater than 1.

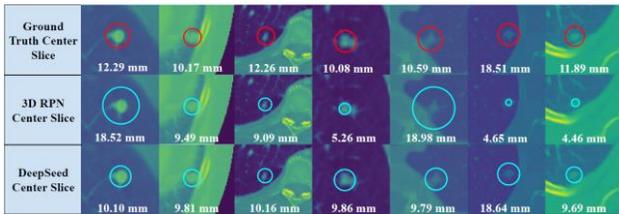

**Fig. 3**. Visualization of representative nodule detection results obtained by the 3D RPN and DeepSEED on LIDC/IDRI scans. Each circle depicts a nodule with its location as the circle's center and its size as the circle's radius.

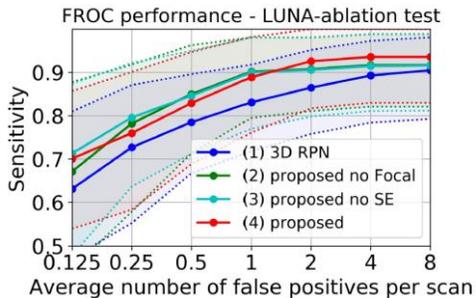

**Fig. 4**. FROC curves obtained by the 3D ResNet-18 backbone and DeepSEED for a randomly selected fold of the cross-validation experiments based on LUNA16 dataset. The dash-lines show the corresponding upper and lower bounding after bootstrapping.

## 4. CONCLUSIONS

In this study, we developed a novel End-to-End 3D deep learning network under an Encoder-Decoder framework with Squeeze-and-Excitation structures for detection of pulmonary nodules in low-dose lung CT scans. To overcome problems caused by sample imbalance and to reduce the false positive rate, we adopted a focal classification loss in the proposed network. Extensive experimental results on LUNA16 and LIDC/IDRI datasets have demonstrated that our model could achieve improved nodule detection performance compared with state-of-the-art methods with the same network backbone. Ablation studies have indicated that both the Squeeze-and-Excitation structure and the focal classification loss were critical to improve the nodule detection performance.

## ACKNOWLEDGEMENTS

This study was supported in part by National Institutes of Health grants [CA223358 and CA189523].